\pdfoutput=1

\documentclass[11pt]{article}

\usepackage[preprint]{acl}

\usepackage{times}
\usepackage{latexsym}

\usepackage[T1]{fontenc}
\usepackage[utf8]{inputenc}
\usepackage{microtype}
\usepackage{inconsolata}
\usepackage{graphicx}
\usepackage{array} 
\usepackage{arydshln} 
\usepackage{amsmath}
\usepackage{multirow} 

\title{Investigating Neurons and Heads in Transformer-based LLMs for Typographical Errors}

\author{
  \textbf{Kohei Tsuji\textsuperscript{1}},
  \textbf{Tatsuya Hiraoka\textsuperscript{2}},
  \textbf{Yuchang Cheng \textsuperscript{1,3}},
  \textbf{Eiji Aramaki \textsuperscript{1}},
  \textbf{Tomoya Iwakura\textsuperscript{1,3}},
\\
\\
  \textsuperscript{1}NAIST,
  \textsuperscript{2}MBZUAI,
  \textsuperscript{3}Fujitsu Ltd.
\\
    tsuji.kohei.tl1@naist.ac.jp,\\ 
    tatsuya.hiraoka@mbzuai.ac.ae,\\ 
    aramaki@is.naist.jp,\\
    \{cheng.yuchang, iwakura.tomoya\}@fujitsu.com,\\ 
}

\begin{document}
\maketitle
\begin{abstract}
This paper investigates how LLMs encode inputs with typos.
We hypothesize that specific neurons and attention heads recognize typos and fix them internally using local and global contexts.
We introduce a method to identify \textbf{typo neurons} and \textbf{typo heads} that work actively when inputs contain typos.
Our experimental results suggest the following: 
1) LLMs can fix typos with local contexts when the typo neurons in either the early or late layers are activated, even if those in the other are not.
2) Typo neurons in the middle layers are responsible for the core of typo-fixing with global contexts.
3) Typo heads fix typos by widely considering the context not focusing on specific tokens.
4) Typo neurons and typo heads work not only for typo-fixing but also for understanding general contexts.

\end{abstract}

\section{Introduction}
Inputs for real applications using large language models (LLMs) sometimes contain typographical errors (typos)~\cite{NoisyExemplars, RobustnessOfChatGPT, Promptrobust}.
LLMs often make correct answers on inputs with typos~\cite{RobustnessOfChatGPT}, which implies that LLMs can ``fix'' typos to recover the initially intended meaning.
However, LLMs sometimes imperfectly fix the meaning against typos, which might ``damage'' the performance of LLMs on downstream tasks~\cite{RobustnessOfCodex, AdvGLUEpp, Promptrobust, CUTE}.
To reduce the impact of typos on LLMs, it is essential to understand both their robustness against typos and the reasons for performance degradation caused by typos more deeply. 

Existing studies have primarily focused on the surface-level exhibition of performance degradation due to typos~\cite{AdvGLUEpp, Promptrobust} and methods for improving robustness against typos~\cite{NoisyExemplars, RobustnessOfCodex, LEA}. 
Few studies have investigated how typos affect LLM's inner workings~\cite{FromTokensToWords, AcronymsTask}.
However, previous work focused on cases where the input contains only a few subwords and a typo. Therefore, they examined typo-fixing working with only local contexts.
In contrast, studies have reported that the performance of typo correction can be improved by observing longer (global) contexts \cite{context_spelling_correction, spellbert}.
This implies that LLMs might see global contexts when handling typo inputs.

Based on these previous works, we hypothesize that LLMs with the Transformer-based decoder also fix typos along two axes: typo-fixing with local contexts, which focuses on nearby subwords, and typo-fixing with global contexts, which understands longer contextual information. 
To verify this hypothesis, we investigated neurons (\textbf{typo neurons}) and attention heads (\textbf{typo heads}) in LLMs that provide robustness against typos through the following steps.
First, we investigated the inner workings against typos in contextualized words using a word identification task (\S \ref{sec:preliminary}).
Then, we propose a method to identify typo neurons (\S \ref{sec:neurons}) and typo heads (\S \ref{sec:head}).
Subsequently, we analyze the differences in their behavior between cases where the model is damaged by typos and cases or not.

We conducted experiments using Gemma 2~\cite{Gemma2}, Qwen 2.5~\cite{qwen2.5}, and two of the Llama 3~\cite{llama3} family to investigate the inner workings when inputs contain typos.
Our findings suggest the following:

\vspace{-0.5\baselineskip}   
\begin{itemize}
    \setlength{\itemsep}{-5pt} 
    \item LLMs can fix typos when the typo neurons in either the early or late layers, both of which focus on local contexts, are activated, even if those in the other are not.
    \item Typo neurons in the middle layers are responsible for typo-fixing considering global contexts, regardless of the models. 
    \item Typo heads fix typos using the local and global contexts, not focusing on specific tokens.
    \item Typo neurons and typo heads not only fix typos but also understand general grammatical or morphological features.
    
\end{itemize}
\begin{figure*}[t]
    \centering
    \includegraphics[width=\linewidth]{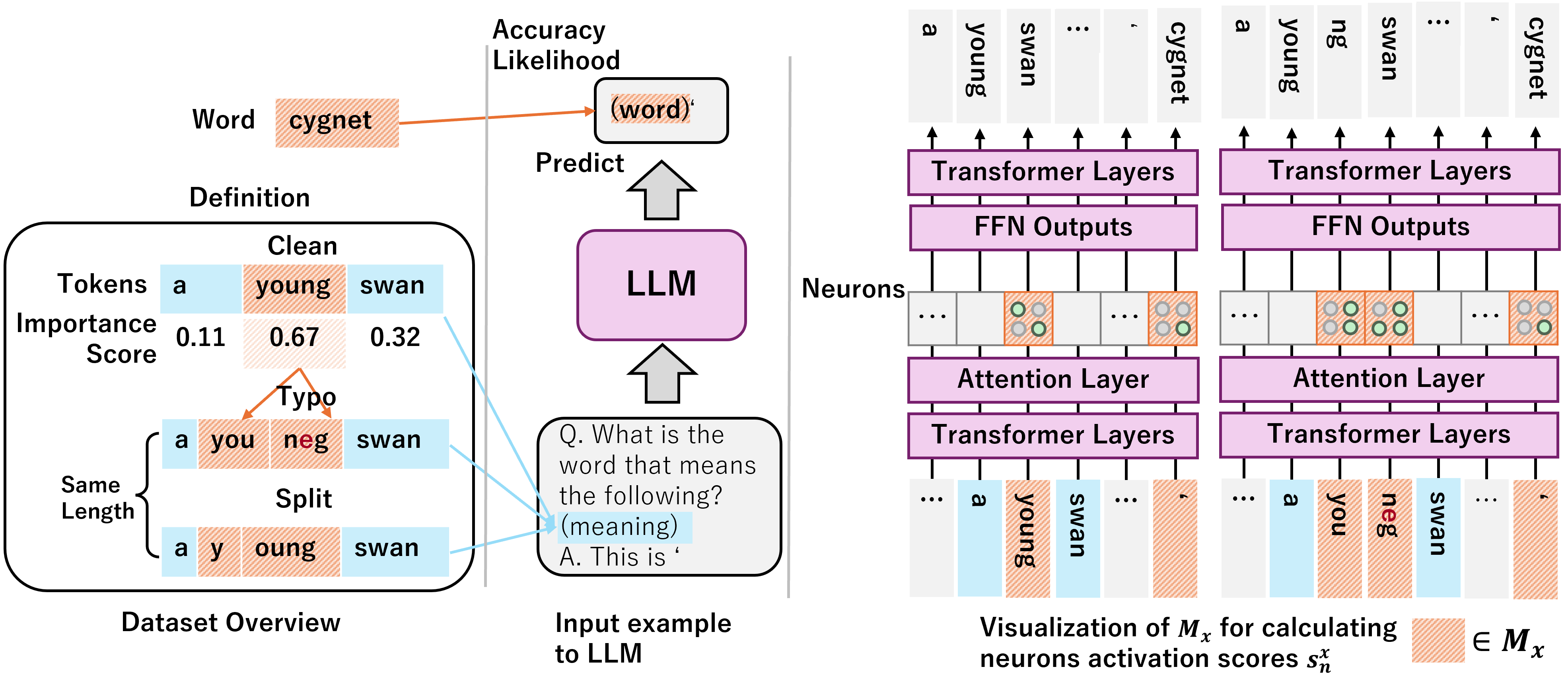}
    \vspace*{-0.7cm}
    \caption{The dataset overview (left), an overview of an input example to LLM (middle), and the visualization of $M_x$ for calculating neurons activation score $s_{n}^{x}$ (right). }
    \label{fig:overview}
\end{figure*}

\section{Related work}
\subsection{Analysis of LLMs against Typos}
Typos are mistakes in writing or typing letters, categorized into insertion, deletion, substitution, and reordering~\cite{DeepWordBug}.
Research on the robustness of LLMs regards typos as a perturbation.
Typos change the token sequence obtained through the tokenization process. 
Changing the token sequence potentially leads to a different output, even if the sentence is the same~\cite{tsuji2024subregweigh}.
Most existing LLM studies about typos focus on measuring the robustness against perturbed inputs~\cite{AdvGLUE, AdvGLUEpp, Promptrobust, CUTE} or modifying the architecture or prompts to improve robustness~\cite{ RobustnessOfCodex, NoisyExemplars, LEA}. 
\newcite{SubwordRobustness} reported that the larger models are more robust to typos. Before the LLM era, researchers corrected typos using specific models for typo-correction~\cite{context_spelling_correction, spellbert}.

\subsection{LLM's Interpretability}
The feed-forward network (FFN) layer in the Transformer~\cite{vaswani2017attention} has two linear layers separated by an activation function.
Recent studies regard the output of the activation function as ``neurons'' that store knowledge~\cite{KVMemory}.
It has been reported that some neurons promote specific tasks~\cite{SkillNeurons, NeuronsAcrossLanguageAndTask}, knowledge~\cite{KnowledgeNeurons, NounPhrasesNeurons, UniversalNeurons}, and behaviors~\cite{RepetitionNeurons, MitigatingRepetition, SafetyNeurons}.

Some attention heads also respond to specific knowledge~\cite{SuccessorHeads, HeadPruning, AcronymsTask} or behaviors~\cite{CopySuppression, CopyInductionHeads}. 
Additionally, some heads are responsible for merging multiple subwords of a word~\cite{SubwordMergeHead, InformationFlowRoutes}.

There are various methods to investigate LLM's interpretability. Some measure contributions to the residual stream~\cite{UnderstandingVulnerabilities, PathPatching}, while others observe intermediate predictions~\cite{LogitLens, FromTokensToWords}, graph the inference process~\cite{InformationFlowRoutes}, or directly observe activations~\cite{SkillNeurons, RepetitionNeurons, NeuronsAcrossLanguageAndTask}.
We hypothesize that typo neurons are a type of skill neurons. Therefore we use the direct activation observation method, following previous studies on skill neurons~\cite{SkillNeurons, RepetitionNeurons}.
\newcite{ImpactOfIA} concludes that understanding the inner workings is important to improve the model performance.

\newcite{StagesOfInference} divides LLMs into four stages. The early layers convert token-level representations into entity-level representations with local contexts as \textit{Detokenization}. The early middle layers build representations with global contexts as \textit{Feature Engineering}. The late middle layers, convert current representations into next token representations as \textit{Prediction Ensembling}. Finally, the late layers remove the noise and refine the distribution of the next token as \textit{Residual Sharpening}.
\newcite{SoftmaxLinearUnits} reports that the late layers perform the opposite function of the early layers' \textit{Detokenization}, converting entity-level representations into token-level representations as \textit{Retokenization}.

\newcite{FromTokensToWords} reveals which layers are responsible for typo-fixing.
However, they only focused on isolated words as inputs by layer-level observation. 
We focus on neurons and heads and experiment with global contexts.

\section{Preliminary} \label{sec:preliminary}
We created a dataset that LLMs can solve without typos (\S \ref{sec:dataset}).
Then, we applied typos to the dataset (\S \ref{sec:injection}) and conducted a preliminary experiment to observe accuracy when inputs include typos (\S \ref{sec:preliminary_experiment}). 
Next, we identify typo neurons and reveal their specific roles (\S \ref{sec:neurons}). 
Similarly, we conduct analogous experiments for attention heads (\S \ref{sec:neurons}).

\subsection{Models}
We used Google's Gemma 2~\cite{Gemma2} with 2B, 9B, and 27B parameters, Meta's Llama 3.2~\cite{llama3} with 1B and 3B parameters, Meta's Llama 3.1 with 8B parameters, and Qwen's Qwen 2.5~\cite{qwen2.5} with 3B, 7B, 14B, 32B parameters; Gemma 2 27B  and Qwen 2.5 32B were loaded in bfloat16, while the others were loaded in float32\footnote{We described our computing environment in Appendix~\ref{sec:computing_environment}.}. We conducted all experiments using greedy generation. 

\subsection{Clean Datasets without Typos} \label{sec:dataset}
We used a word identification task in which LLMs infer a single word from a given definition. Since typo-fixing relies on vocabulary knowledge, it is crucial to use a task that directly reflects the LLMs' vocabulary knowledge, such as word identification. Moreover, we avoided tasks requiring complex reasoning, such as NLI, as variations in sample difficulty could hinder a clear observation of typo-related phenomena.

For instance, we feed the definition of the word as input, like ``\textit{a young swan}'', to an LLM, and then the model is expected to output the corresponding word ``\textit{cygnet}''.
Following \newcite{VerbWordNet}, we extracted 62,643 word-definition pairs from WordNet~\cite{WordNet}\footnote{WordNet via NLTK~\cite{bird-loper-2004-nltk} ver.3.9.1.}. 
We created the dataset with these pairs. 
We designed a prompt so that LLMs can solve this task by predicting tokens following outputs, as shown in the middle part of Figure~\ref{fig:overview}.

For our analysis, we need a dataset composed of samples that LLMs can correctly answer when the samples do not include typos.
Therefore, we extracted the top 5,000 or 1,000 word-definition pairs after sorting the samples by descending order of likelihood for the correct words\footnote{Due to Llama 3.2 1B's worse performance, we could not extract 5,000 pairs for the Llama 3 family. Therefore, we extract 1,000 pairs for the Llama 3 family.}.
Note that we created a unique dataset for each model. 

\subsection{Generating Inputs with Typos} \label{sec:injection}
\subsubsection{Typo Dataset} \label{sec:typo_dataset}
To focus on text with typos, we generated inputs with typos from the definition part of the clean dataset created in \S \ref{sec:dataset}.
We selected the top $t$ most important tokens depending on their importance scores on the word identification task.
Then, we injected a random single letter or digit into each selected token as a typo.
The importance scores were calculated with the method used in \newcite{AdvGLUEpp, TextBugger}, with the smallest models among ones that share the same tokenizer (e.g., Gemma 2 2B for Gemma 2 or Llama 3.2 1B for Llama 3 family). 
Specifically, we obtained the importance scores by performing back-propagation while predicting words from their definitions.
This process assigns higher gradients to tokens that are important to predict the correct answer.
For example, consider the sentence ``\textit{a young swan}'' with $t=2$ and the top two most important words are ``\textit{young}'' and ``\textit{swan}.''
In this case, we inject random letters such as ``\textit{e}'' and ``\textit{5}'' into random positions\footnote{We exclude the positions before the spaces to avoid the situation where a typo would appear at the end of the previous token rather than within the target token.} of each word, which results in ``\textit{a youn\textbf{e}g s\textbf{5}wan}.''

\subsubsection{Split Dataset}
We often obtain a different number of subwords when tokenizing typo inputs compared to clean inputs.
For instance, the tokenizer encodes the word ``young'' into a single token, but it tokenizes the typo version ``youneg'' into two tokens (e.g., ``you / neg'').
When comparing the inner workings when LLMs encode the clean inputs and the typo inputs, the difference in the token length might prevent appropriate analysis\footnote{\newcite{FromTokensToWords} reported that there are inner workings to fix the original token from differently tokenized subwords. We need to exclude the effect of this factor to deeply focus on the typo-related inner workings.}.

To divide typo-related inner workings into the factor corresponding to typos and the one to tokenization difference, we created the ``split dataset'' in addition to the ``typo dataset'' mentioned in \S \ref{sec:typo_dataset}.
The split dataset contains samples tokenized into the same number of tokens as the one with typos.
For example, when the typo dataset has a sample whose tokenized sequence is ``a / you / neg / swan'', an example of counterparts in the split dataset is ``a / y / oung / swan'' whose length is equivalent to the one of the typo version.
We can obtain the various tokenization candidates using the tokenizer and we randomly selected one candidate with the same length as the typo input.
This process is shown in Figure~\ref{fig:overview} (left).

\begin{figure}[t]
    \centering
    \includegraphics[width=\linewidth]{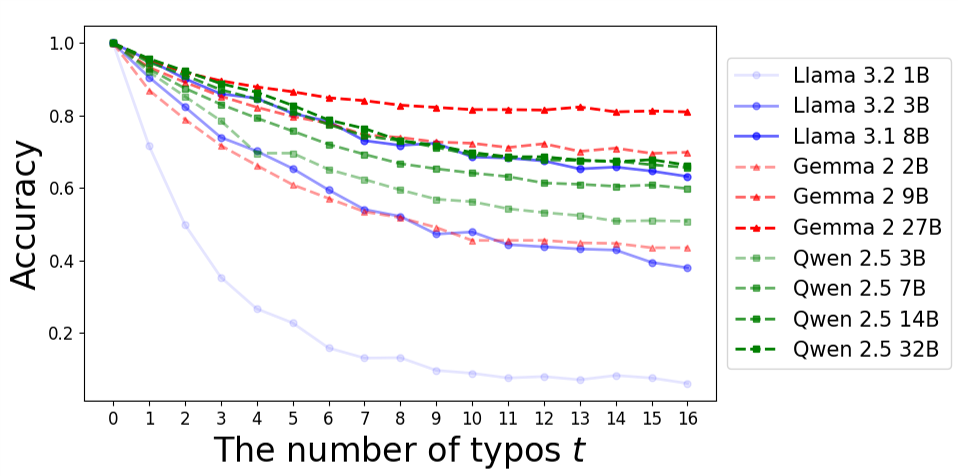}
    \caption{Accuracy on the word identification task with different numbers of typos $t$.}
    \label{fig:acc}
\end{figure}

\begin{figure*}[t]
    \centering
    \includegraphics[width=\linewidth]{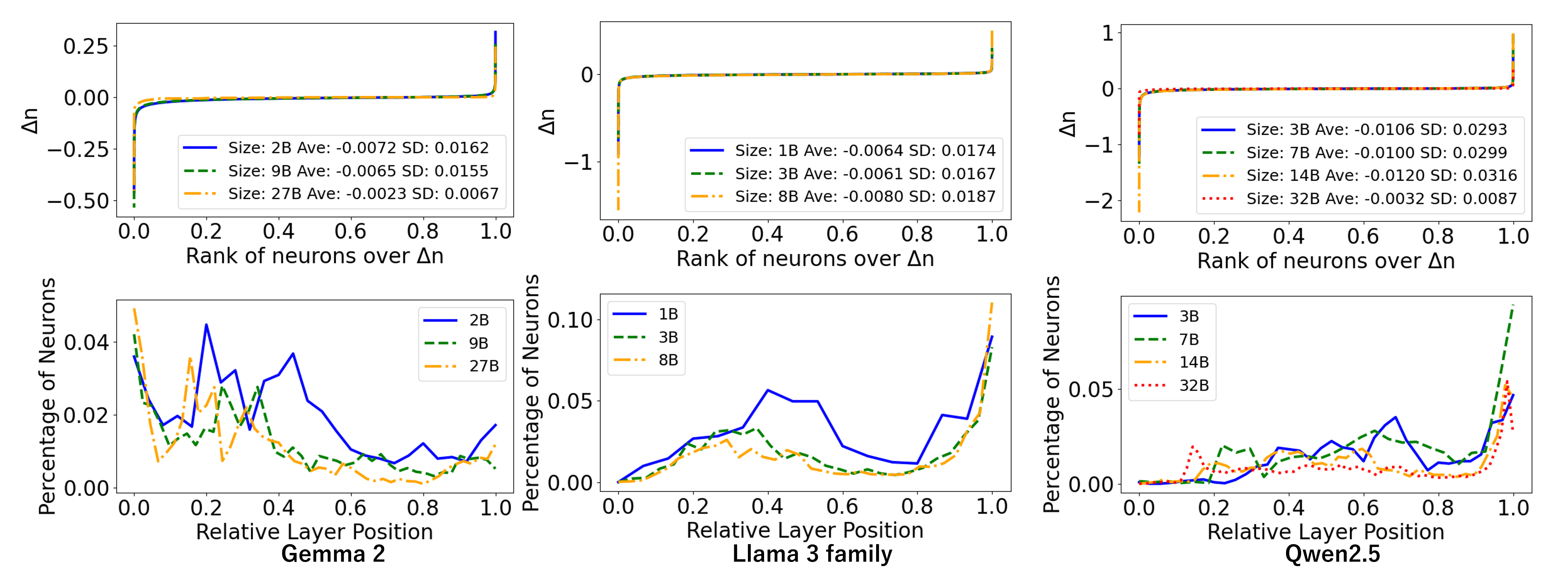}
    \caption{Distribution of ${\Delta}_{n}$ (upper) and percentage of typo neurons per layer (lower) with $t=1$. The left figures are for Gemma 2, the center figures are for Llama 3 family and the right figures are for Qwen 2.5.} 
    \label{fig:neurons_results}
\end{figure*}

\subsection{Preliminary Experiment} 
\label{sec:preliminary_experiment}
To examine the impact of typos on the model performance, we applied typos to $t$ tokens ($ 1 \leq t \leq 16 $) and analyzed the change in accuracy.

Figure~\ref{fig:acc} shows the preliminary experimental results. 
The accuracy of $t=0$ indicates the performance of the clean data without typos. 
Since the clean data consists of samples that each model can answer correctly, the accuracy for all models is 1.0. 
As shown in the figure, the larger models maintain higher accuracy than the smaller models even with many typos.
This result supports the existing work reporting that the larger model has robustness against typos~\cite{SubwordRobustness}.
This preliminary result also indicates that the robustness of larger models against typos is insufficient, resulting in a performance drop.
We conclude that typos damage performance, but larger LLMs have some robustness against typos, which motivates us to investigate the typo-related inner workings.
Furthermore, this leads us to a deep analysis of the reasons for the differences in robustness against typos by model sizes for further improvement.

\section{Typo Neurons}
\label{sec:neurons}
Some FFN layers have been found to combine multiple tokens into a single representation vector~\cite{FromTokensToWords, SoftmaxLinearUnits, StagesOfInference}.
Additionally, it has been reported that certain neurons within LLMs function as ``skill neurons'' with specific roles~\cite{SkillNeurons}.
In this section, we investigate the existence of typo neurons, a particular type of skill neuron that is responsible for recognizing and fixing typos.

\subsection{Method to Identify Typo Neurons}
\label{sec:neuron_method}
Following the approach of \newcite{RepetitionNeurons}, we compare the activation values of neurons between clean inputs and typo inputs to identify neurons that specifically respond to typos.
Let $x \in X$ be a sample of the dataset, where $x$ is a sequence of $|x|$ tokens: $x={w_1, ..., w_m, ..., w_{|x|}}$. Each sample comprises the prompt (e.g., ``\textit{Q. What is ... A. This is }'') and the answer (e.g., ``\textit{cygnet}'').

The activation value $s_n^X$ of a neuron $n$ when feeding a dataset $X$ is defined as the following:
\begin{equation}
s_n^X = \frac{1}{|X|}\sum_{x\in X} \left(\frac{1}{|M_x|}\sum_{m \in M_x} f(x_1^m,n)\right),
\end{equation}
where $|X|$ is the number of samples in the dataset.
$f(x_1^{m}, n)$ is a function calculating the activation value of the neuron $n$ corresponding to $w_m$ when the LLM reads the input $x_1^m = {w_1, ..., w_m}$.
$M_x$ is a set of indices that indicates the token positions, and $|M_x|$ is the number of indices.
We define $M_x$ as the indices comprising the answer word tokens and $t$ important words.

For example, in Figure \ref{fig:overview}, $M_x$ for the clean input is composed of ``young'' and the apostrophe before ``cygnet'', while $M_x$ for the typo input is composed of ``you'', ``neg'', and the apostrophe and for the split input is ``y'', ``oung'', and the apostrophe.
In the figure, tokens comprising $M_x$ are indicated with an orange background.

We obtain the responsibility of neurons specialized to the typo inputs separated from clean and split inputs with the following score $\Delta_n$:
\begin{equation}
\label{eq:delta_n}
{\Delta}_{n} = s_n^{X_{\mathrm{typo}}} - \max\left(s
_n^{X_{\mathrm{clean}}},s_n^{X_{\mathrm{split}}}\right),
\end{equation}
where $X_{\mathrm{typo}}$, $X_{\mathrm{clean}}$, and $X_{\mathrm{split}}$ are the typo, clean, and the split datasets, respectively. 

A larger ${\Delta}_{n}$ indicates the neuron $n$ that responds specifically to typos but not clean inputs or split inputs. 
Among the neurons, the top $K$ neurons based on ${\Delta}_{n}$ scores are identified as typo neurons.

\subsection{Experimental Results}
\label{sec:neuron_results}
This section investigates the typo neurons found with the method introduced in \S \ref{sec:neuron_method}.
We used the number of typos $t = 1$. 
Appendix~\ref{sec:t_16_neurons} additionally describes the results for $t=16$.

Figure \ref{fig:neurons_results} shows the distribution of ${\Delta}_{n}$ and the distribution of the typo neurons in each layer.
We extracted the top $0.5$\% of neurons with the highest ${\Delta}_{n}$ as the typo neurons. 
The average (Ave) and standard deviation (SD) in Figure \ref{fig:neurons_results} indicate that a few neurons have significantly larger scores than others, similar to knowledge and skill neurons~\cite{KnowledgeNeurons, SkillNeurons}.

For the distribution of neurons, Llama 3 family and Qwen 2.5 have many typo neurons in the late layers(i.e., from 0.8 to 1.0). 
In contrast, Gemma 2 models have many typo neurons in the early layers (i.e., from 0.0 to 0.2), and few are in the late layers. 
Especially in the 9B and 27B models, the largest number of typo neurons exist in the early layers. 

According to \newcite{StagesOfInference}, the late layers perform \textit{Residual Sharpening}, which removes noise from representations. Considering typos as noise, it is natural that many typo neurons are in the late layers.
Besides, \newcite{SoftmaxLinearUnits} reports that the early layers are responsible for \textit{Detokenization} that converts raw token representations into coherent entities (e.g., words), while the late layers perform \textit{Retokenization} that converts them back into token-level representations. 
These suggest that Gemma 2 fixes typos as \textit{Detokenization}, while LLaMA 3 family and Qwen 2.5 fix typos as \textit{Retokenization}.
Since both processes use local contexts, we can see the variety of the balance in responsibility between the early and late layers. 
As shown in Appendix~\ref{sec:t_16_neurons}, with many typos, typo neurons in the late layers of Gemma 2 models also increased. This indicates that the distribution of responsibility between the early and late layers is adaptable.

In the middle layers (i.e., 0.2-0.8), all models have many typo neurons. This suggests that these layers play a common role in typo-fixing across models. 
Since the early middle layers create representations depending on global contexts with attention heads as \textit{Feature Engineering} and the late middle layers convert current representations to next token representations as \textit{Prediction Ensembling}~\cite{StagesOfInference}, typo-fixing in these layers seem to focus on recognition of global contexts in contrast to the early and late layers.

\subsection{Discussion}
\label{sec:neuron_discussion}
While the experimental results in \S \ref{sec:neuron_results} suggest the existence of typo neurons, their impact has not been clarified. 
Then, in this section, we investigate their specific impact, focusing primarily on Gemma 2. 

\subsubsection{Neuron ablation}
\label{sec:neuron_ablation}
\begin{table}[t]
    \small
    \centering
    \begin{tabular}{ll|rr}
        \hline \hline
         &  & \multicolumn{1}{l}{\begin{tabular}[c]{@{}l@{}}Clean\\ Dataset\end{tabular}} & \multicolumn{1}{l}{\begin{tabular}[c]{@{}l@{}}Typo\\ Dataset\end{tabular}} \\ \hline
        \multicolumn{2}{l|}{Gemma 2 2B} & 1.00 & 0.86 \\
         & $\ominus$ Random Neurons& 0.98 & 0.87 \\
         & $\ominus$ Typo Neurons& 0.84 & 0.73 \\ \hdashline
        \multicolumn{2}{l|}{Gemma 2 9B} & 1.00 & 0.93 \\
         & $\ominus$ Random Neurons & 0.99 & 0.96 \\
         & $\ominus$ Typo Neurons & 0.93 & 0.90 \\ \hdashline
        \multicolumn{2}{l|}{Gemma 2 27B} & 1.00 & 0.95 \\
         & $\ominus$ Random Neurons & 0.98 & 0.94 \\
         & $\ominus$ Typo Neurons & 0.96 & 0.91 \\ \hline \hline
    \end{tabular}
    \caption{Accuracy of the word identification task with neuron ablation on clean and typo datasets. 
    ``$\ominus$ Random/Typo Neurons'' indicates the performance by ablating random and typo neurons, respectively.
    }
    \label{tab:deactivate_result}
\end{table}
We expect typo neurons to work typo-fixing.
Therefore, ablating them should result in a remarkable decrease in performance for typo inputs while not affecting the performance for clean inputs.

We test this hypothesis by conducting ablation experiments on typo neurons and randomly selected neurons of Gemma 2 models.
Appendix~\ref{sec:neuron_ablation_others} discusses the results of the ablation study for other models.
From a dataset of 5,000 samples, 100 randomly selected samples were used to identify typo neurons.
Then, we evaluate the performance of the word identification task using the remaining 4,900 samples by deactivating the identified neurons. 

Following \S \ref{sec:neuron_results}, we identified the top 0.5\% of neurons as typo neurons.
We also randomly selected 0.5\% of neurons as a baseline.
Deactivation was performed by setting the output values of the neurons to zero.
The experiments were conducted for the clean inputs and the typo inputs with $t=1$.

Table~\ref{tab:deactivate_result} shows the experimental results.
For typo inputs, performance remained largely unchanged when random neurons were ablated, regardless of the model.
However, performance decreased when typo neurons were ablated. 
This suggests that a small number of typo neurons play an important role in typo-fixing for typo inputs.
For clean datasets, the ablation of typo neurons also resulted in a larger performance decrease than the random neuron ablation. 
This indicates that typo neurons may not exclusively act on typos but could also play a crucial role in processing general grammatical or morphological features.
We can see similar results with the other models (Appendix~\ref{sec:neuron_ablation_others}).

\subsubsection{Neurons for Typo-fixing}
\begin{figure}[t]
    \centering
    \includegraphics[width=\linewidth]{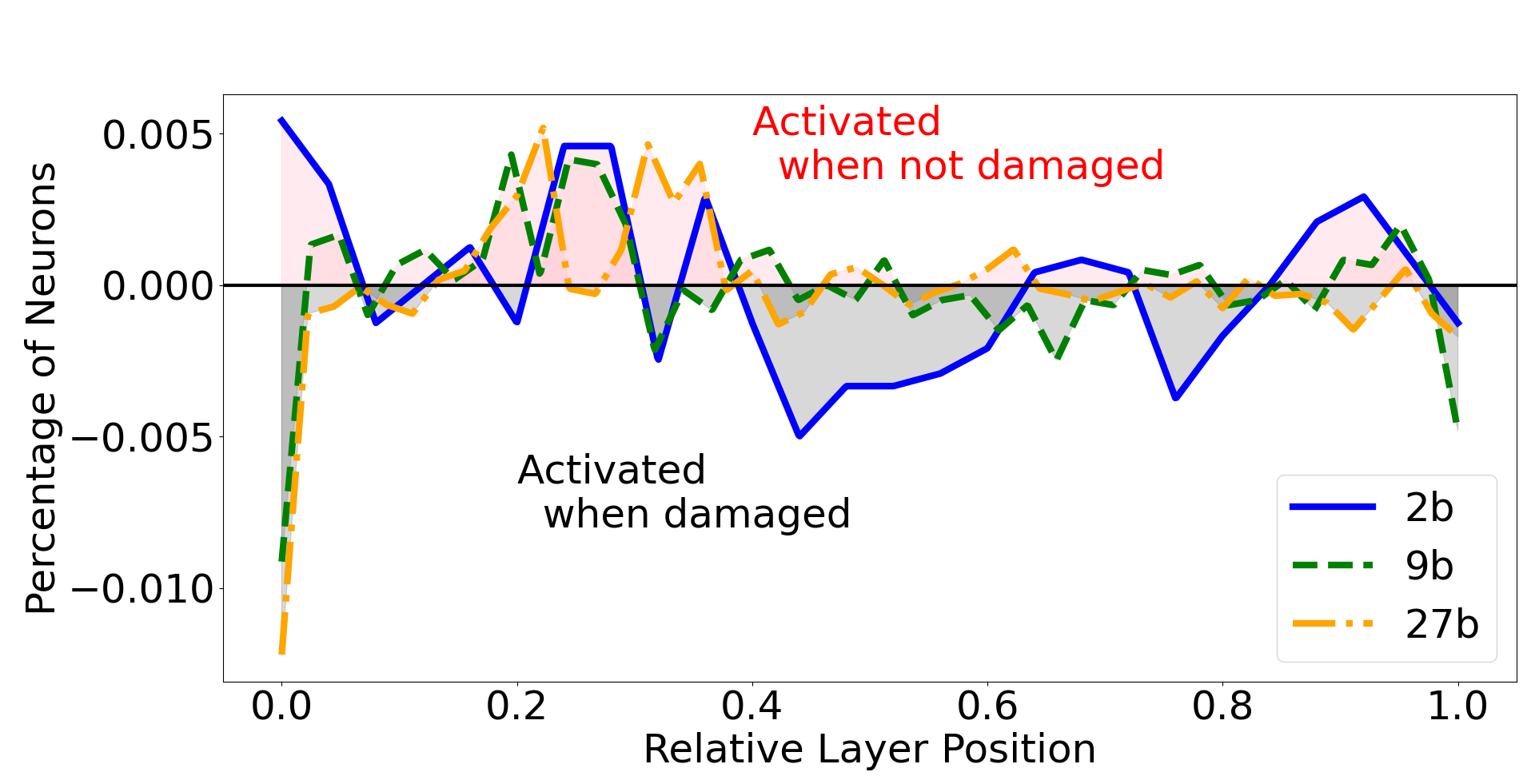}
    \caption{Distribution of typo neurons per layer for samples damaged or not. 
    Values above the black line indicate many typo neurons activated when the LLMs predicted correct words.}
    \label{fig:damaged_or_not}
\end{figure}
The experiments in \S \ref{sec:neuron_results} sought typo neurons by comparing clean and typo inputs without considering whether the LLMs could correctly solve the task with typo inputs.
This section focuses on the difference in typo neurons between cases where the LLMs answer with typos correctly and incorrectly.

From the dataset of 5,000 samples, we extracted 100 samples where typos did not damage the inferences and the correct word was predicted.
Similarly, we extracted another 100 samples where typos damaged the inferences and led to incorrect word prediction. 
We compared differences in the activation of typo neurons in these two groups.
We conducted this experiment with $t=1$ and compared the difference in the layer distribution of the typo neurons that have the top 0.5\% ${\Delta}_{n}$.

Figure~\ref{fig:damaged_or_not} shows the result.
In the 9B and 27B models, the number of typo neurons in the early layers increases when incorrect inferences are predicted. 
This suggests that some neurons in the early layers might play other roles than typo-related phenomena, and activation of those neurons prevents correct recognition of typos.
In the 2B model, when the model fails to fix typos, typo neurons in the middle-middle layers are activated. This suggests that the strong activations observed in the middle-middle layers of Gemma 2 2B in \S \ref{sec:neuron_results} are due to neurons damaged by typos rather than contributing to typo-fixing.
Across all models, more typo neurons in the early middle layer (e.g., 0.2-0.4) were activated when typos did not damage inferences. This indicates the importance of typo neurons in the early middle layers.

\begin{figure*}[t]
    \centering
    \includegraphics[width=\linewidth]{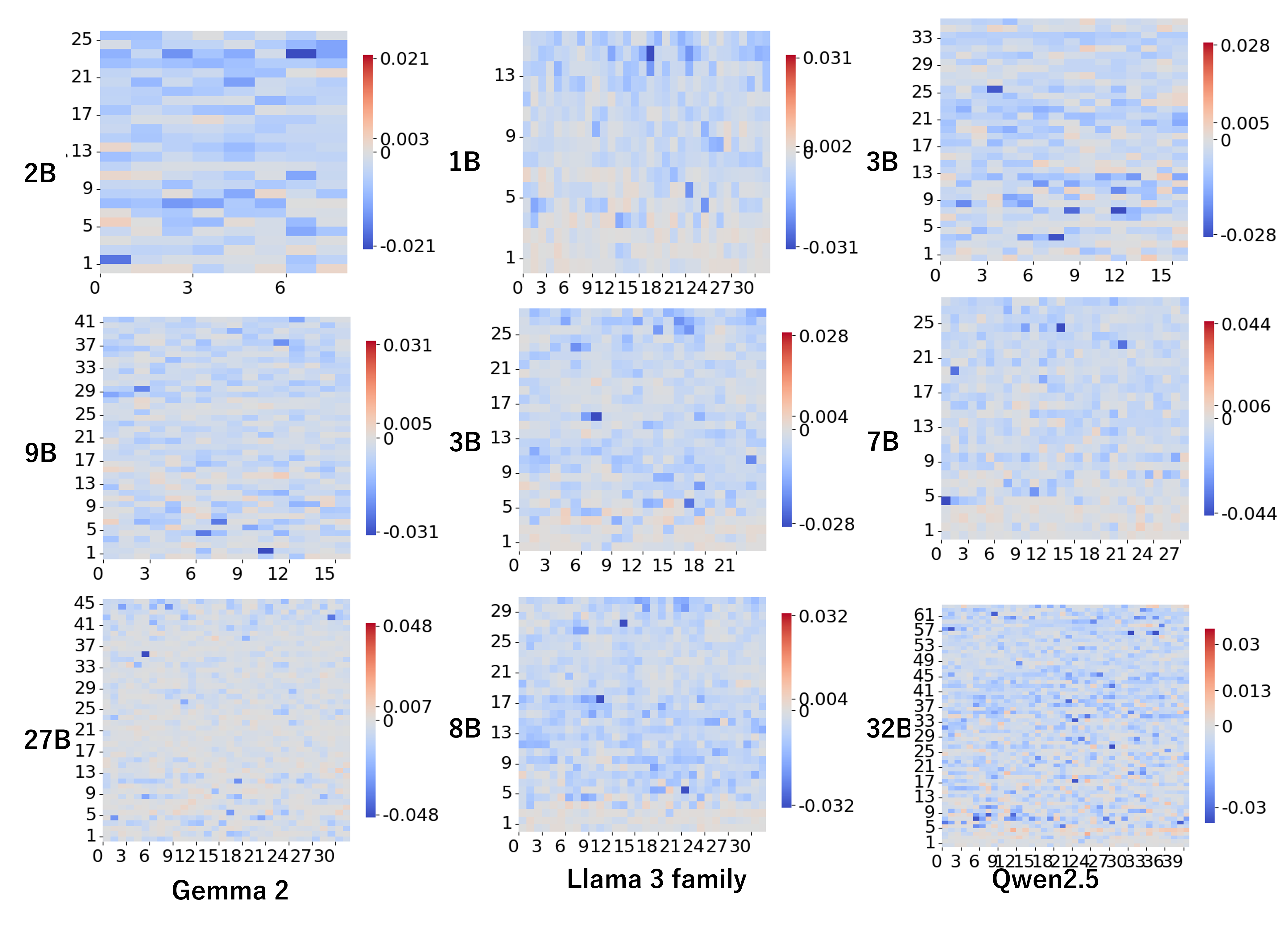}
    \vspace*{-0.8cm}
    \caption{Distribution of ${\Delta}_{h}$ for each model with $t=1$. 
    The heat map colors are centered around 0, and the tick mark closest to 0 on the positive side of the heat bar represents the maximum ${\Delta}_{h}$. The left figures are for Gemma 2, the center figures are for Llama 3 family and the right figures are for Qwen 2.5.}
    \label{fig:head_delta}
\end{figure*}

\begin{table*}[t]
    \small
    \centering
    \begin{tabular}{l|rrr:rrr:rrrr}
        \hline \hline
         & \multicolumn{3}{c:}{Gemma 2} & \multicolumn{2}{c}{Llama 3.2} & \multicolumn{1}{c:}{Llama 3.1} & \multicolumn{4}{c}{Qwen 2.5} \\ 
         & \multicolumn{1}{c}{2B} & \multicolumn{1}{c}{9B} & \multicolumn{1}{c}{27B} & \multicolumn{1}{c}{1B} & \multicolumn{1}{c}{3B} & \multicolumn{1}{c}{8B} & \multicolumn{1}{c}{3B} & \multicolumn{1}{c}{7B} & \multicolumn{1}{c}{14B} & \multicolumn{1}{c}{32B} \\ \hline
        Average & -0.0045 & -0.0042 & -0.0032 & -0.0040 & -0.0039 & -0.0049 & -0.0043 & -0.0053 & -0.0047 & -0.0050 \\
        \begin{tabular}[c]{@{}l@{}}SD \end{tabular} & 0.0038 & 0.0041 & 0.0049 & 0.0045 & 0.0040 & 0.0044 & 0.0046 & 0.0056 & 0.0052 & 0.0057 \\ \hline \hline
    \end{tabular}
    \caption{The average and standard deviation (SD) of $\Delta_{h}$.}
    \label{tab:head_ave_sd_t1}
\end{table*}

\section{Typo Heads}
\label{sec:head}

\subsection{Method to Identify Typo Heads}
Typo-fixing may not solely depend on neurons but subword merging by attention heads~\cite{SubwordMergeHead, InformationFlowRoutes} and is based on understanding local and global contexts. 
We assume two types of such heads for typo inputs: 1) the one focusing on important tokens and 2) the one widely attending contexts.

In this section, we investigate the attention heads specialized to typo inputs by comparing attention maps.
Herein, we calculated the KL divergence between a uniform distribution and the rows of attention maps by considering them as a probability distribution.
The KL divergence increases monotonically with the number of tokens, which can result in higher values for typo inputs or split inputs, as they often have more tokens than clean inputs. 
We alleviate this problem by normalizing the KL divergence with the maximum score $\log_2 m$, defined as follows:
\begin{equation}
s_h^{X} = \frac{1}{|X|}\sum_{x\in X}\left(\sum_{m}\left(\frac{D_{\mathrm{KL}}(P_{x,m,h}||U_m)}{\log_2 m} \right)\right),
\end{equation}
where $D_{\mathrm{KL}}(\cdot)$ is the function that returns the KL divergence, $U_m$ is a uniform distribution over $m$ elements.
$P_{x,m,h}$ is the $m$-th row of the attention map output by head $h$ for the token sequence $x$. 
In decoder models, attention scores for the $m$-th token and each token from the first to the $m$-th token sum to 1. 
Unlike neurons, for the calculation of typo head identification, we did not narrow down the tokens to calculate and used all tokens in prompts.

Similar to Eq.~\eqref{eq:delta_n} in neurons, the responsibility score of the heads to the typos is defined as follows:
\begin{equation}
{\Delta}_{h} = s_h^{X_{\mathrm{typo}}} - \max\left(s_h^{X_{\mathrm{clean}}}, s_h^{X_{\mathrm{split}}}\right),
\end{equation}
where $X_\mathrm{typo}$, $X_\mathrm{clean}$, and $X_\mathrm{split}$ are the typo, clean, and split datasets, respectively. 
A large absolute value of ${\Delta}_{h}$ indicates that the head behaves much differently for typo inputs than for clean ones. 
Specifically, a large positive ${\Delta}_{h}$ indicates the head that focuses on specific tokens for typo-fixing, while a large negative ${\Delta}_{h}$ indicates the head that widely attends contexts for typo-fixing.
We identified the top $J$ heads with the highest absolute value of ${\Delta}_{h}$ as typo heads.

\subsection{Experimental Results}
\label{sec:head_results}
We used the number of typos $t = 1$. 
Appendices~\ref{sec:t_16_heads} and \ref{sec:qwen14b_heads} discuss other settings.
As shown in Figure~\ref{fig:head_delta}, the differences between the maximum and absolute minimum scores are approximately 10 times in all models. 
The average and standard deviation in Table~\ref{tab:head_ave_sd_t1} also indicate that few heads near the minimum $\Delta_{h}$ are distinctive.
These results suggest that heads recognize and fix typos by observing the wider context, not by focusing on specific tokens.

As the model size increases, the proportion of heads with ${\Delta}{h}$ close to zero increases. 
This contrasts with the results in \S \ref{sec:neuron_results}, where model differences contributed to the difference in the distribution of typo neurons.
However, we can see a similar trend between the distributions of typo neurons and typo heads in very early layers ($\sim 10\%$ layers from the first layer).
For instance, Gemma 2 has some heads with large ${\Delta}{h}$ in these layers while the Llama3 family and Qwen 2.5 do not.
This trend among models is similar to the one in the distribution of typo neurons (see Figure \ref{fig:neurons_results}).

\subsection{Discussion}
\label{sec:head_discussion}
In this section, we investigate the specific impact and behavior of typo heads, focusing primarily on Gemma 2 similar to \S \ref{sec:neuron_discussion}.

\subsubsection{Head Ablation}
\label{sec:head_ablation}
\begin{table}[t]
    \small
    \centering
    \begin{tabular}{ll|rr}
        \hline \hline
         &  & \multicolumn{1}{l}{\begin{tabular}[c]{@{}l@{}}Clean\\ Dataset\end{tabular}} & \multicolumn{1}{l}{\begin{tabular}[c]{@{}l@{}}Typo\\ Dataset\end{tabular}} \\ \hline
        \multicolumn{2}{l|}{Gemma 2 2B} & 1.00 & 0.86 \\
         & $\ominus$ Random Heads& 0.87 & 0.80 \\
         & $\ominus$ Typo Heads& 0.81 & 0.75 \\ \hdashline
        \multicolumn{2}{l|}{Gemma 2 9B} & 1.00 & 0.93 \\
         & $\ominus$ Random Heads & 0.80 & 0.76 \\
         & $\ominus$ Typo Heads & 0.89 & 0.81 \\ \hdashline
        \multicolumn{2}{l|}{Gemma 2 27B} & 1.00 & 0.95 \\
         & $\ominus$ Random Heads & 0.35 & 0.33 \\
         & $\ominus$ Typo Heads & 0.69 & 0.67 \\ \hline \hline
    \end{tabular}
    \caption{Accuracy of the word identification task with head ablation on clean and typo datasets. 
    ``$\ominus$ Random Heads'' and ``$\ominus$ Typo Heads'' indicate the performance by ablating random and typo heads, respectively.
    }
    \label{tab:head_deactivate_result}
\end{table}

Following the approach in \S \ref{sec:neuron_ablation}, we identified typo heads in Gemma 2 using 100 randomly selected samples of the dataset. 
Then, we ablated these identified typo heads and measured the accuracy on the remaining 4,900 samples. 
Since the total number of heads is smaller than neurons, we identified the top 1.5\% of heads as typo heads (e.g., $J = {3,10,22}$ for 2B, 9B, 27B, respectively). 
We also randomly selected 1.5\% of heads as a baseline.
We performed ablation by setting all attention scores of the selected heads to 0. 
The experiments were conducted for the clean inputs and the typo inputs with $t=1$. We described the results of the ablation study for other models in Appendix~\ref{sec:head_ablation_others}.

Table \ref{tab:head_deactivate_result} shows the experimental result.
In the 9B and 27B models, the ablation of random heads significantly damages the performance in both clean and typo datasets compared to the typo heads, while the ablation of typo heads also degrades the performance to some degree.
This suggests that typo heads are less important in typo-fixing than other heads, while typo neurons have an important role for both typo and clean inputs in \S \ref{sec:neuron_ablation}.
In contrast, for the 2B model, which has fewer heads, the ablation of typo heads resulted in a greater decrease in accuracy than the ablation of random heads.
This suggests that when the number of heads and parameters are limited, they are actively used for typo-fixing. 

In summary, the importance of typo heads is minor in larger models but higher in smaller models. Additionally, since the ablation of typo heads also reduces accuracy on clean datasets, typo heads may play a role in processing general contextual information like typo neurons.

\section{Conclusion}
This paper investigated how the neurons and heads of Transformer-based LLMs respond to typo inputs.
Experimental results show that LLMs can fix typos with local contexts when the typo neurons in either the early or late layers are activated even if those in the other are not. 
While they fix typos by recognizing local contexts, typo neurons in the middle layer are responsible for the core of typo-fixing with global contexts.
Typo heads fix typos using the context widely rather than focusing on specific tokens. Additionally, typo heads are more critical for smaller models than for larger models.

Our findings indicate that Transformer-based LLMs fix typos with not only local but also global contexts, which suggests that improving typo robustness requires approaches that emphasize recognition of both local and global contexts.
The results of the ablation study show that typo-fixing is related to general grammatical or morphological recognition, which suggests that methods for improving typo robustness may also enhance general contextual recognition performance. These findings also suggest that aiming at improving general contextual recognition could contribute to typo robustness.

\section*{Limitation}
This work focuses on the investigation of typo-related inner workings. We believe our findings will help develop applications to alleviate the performance decrease caused by typo inputs. However, the discussion of a concrete method for this application is out of the scope of this paper.
Our analysis was limited to Gemma 2, Llama 3 family, and Qwen 2.5 models and examined models with sizes up to 32B. Larger models or LLMs with different architectures may have different properties. 
For hyperparameters, our experiments were performed only at $t \in \{1,16\}$. 
Furthermore, our experiments focused on a specific task, and models may show different properties in a wider variety of tasks. 
We ran all experiments only once, although there was randomness in applying typos and conducting some experiments. 
For typo neurons, models were observed to have either more typo neurons in the early layers or more in the late layers. This may be due to differences in training methods or datasets. However, the true reason remains unclear.
Additionally, our method mostly detected neurons and heads that respond to inputs with typos. However, it cannot distinguish between those that contribute to typo-fixing and those that are damaged by typos.

\bibliography{main}

\newpage
\appendix

\section{Computing Environment}
\label{sec:computing_environment}
We used NVIDIA A100 40GB$\times$2 for Gemma 2 and Llama 3.1 8B, NVIDIA A100 80GB$\times$1 for Qwen 2.5, and NVIDIA RTX 3060$\times$1 for Llama 3.1 1B and 3B. 

\section{Models Using the Same Tokenizer}
Since LLMs using the same tokenizer share their vocabulary, the impact of typos could be similar. To compare LLMs using the same tokenizer under similar settings, we constructed datasets for such models so that they contain as many identical samples as possible. 

\section{Typo Neurons for Many Typos}
\label{sec:t_16_neurons}
\begin{figure*}[t]
    \centering
    \includegraphics[width=\linewidth]{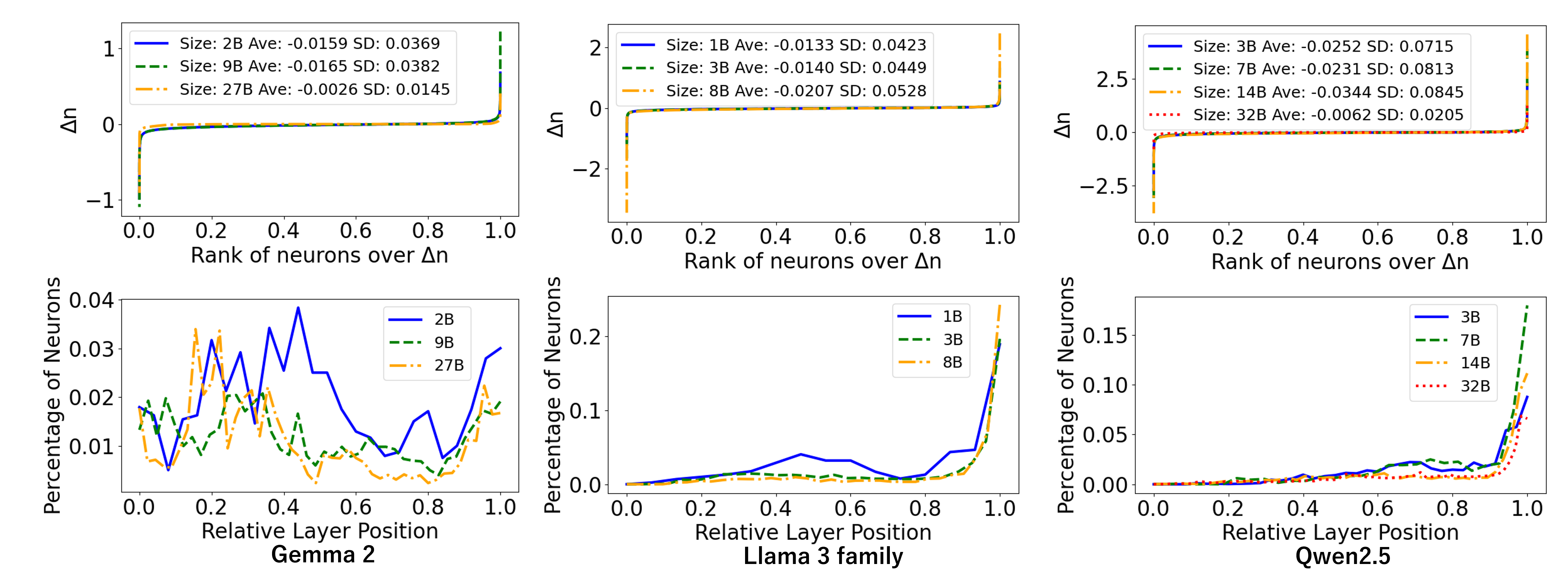}
    \caption{Distribution of ${\Delta}_{n}$ (upper) and percentage of typo neurons per layer (lower) with $t=16$. The left figures are for Gemma 2, the center figures are for Llama 3 family and the right figures are for Qwen 2.5.} 
    \label{fig:neurons_results_t16}
\end{figure*}

\begin{figure*}[t]
    \centering
    \includegraphics[width=\linewidth]{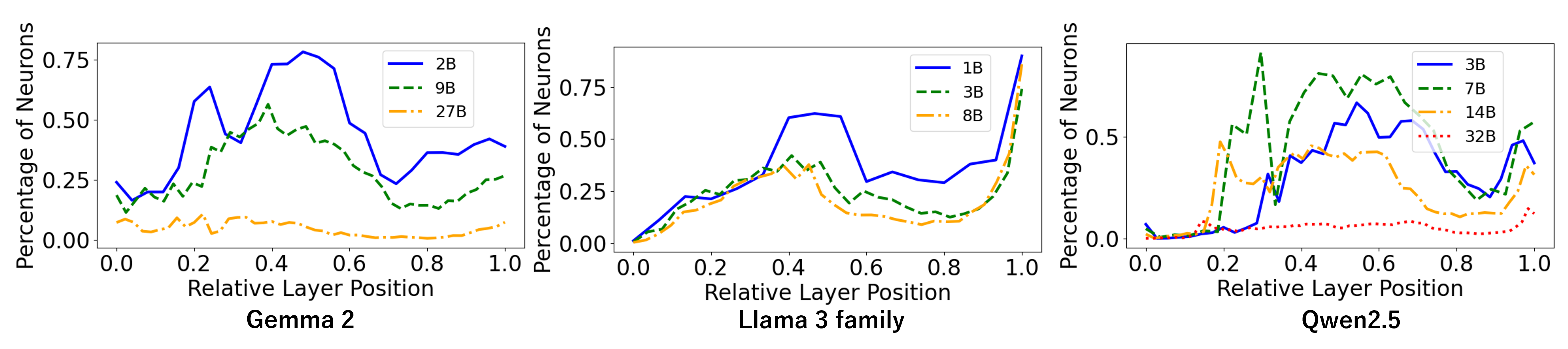}
    \caption{Percentage of typo neurons per layer with $t=16$ when we extracted neurons that activate greater than the typo neurons at $t=1$ as typo neurons. The left figures are for Llama 3 family, the center figures are for Qwen 2.5 and the right figures are for Gemma 2.} 
    \label{fig:neurons_results_t16_diff}
\end{figure*}

In \S \ref{sec:neuron_results}, we reported the results for $t=1$. Here, we describe the behavior of typo neurons with $t=16$, where many typos are introduced. 
Since we are comparing $t=1$, which contains a minimal number of typos, with $t=16$, which has an unrealistically high number of typos, it is expected that the behavior for real-world typos would fall somewhere between them.

Figure~\ref{fig:neurons_results_t16} (upper) shows that the maximum value of $\Delta_{n}$ increases across all models. This indicates that typo neurons respond more strongly as the number of typos increases.
Since the average and standard deviation remain close to zero, it suggests that even in such environments, most neurons activate similarly to those with clean input.

For the Llama 3 family and Qwen 2.5, the proportion of typo neurons in the late layers increases further, while there are few typo neurons in other layers.
However, We extracted only the top 0.5\% of neurons with the highest $\Delta_{n}$ as the typo neurons.
Therefore, even if neurons in other layers are activated similarly to those in $t=1$, a significant increase in typo neuron activation in the late layers could cause a ranking inversion of $\Delta_{n}$. This leads to the possibility that some activated neurons are not extracted as the typo neurons.

To address this, we redefine typo neurons by extracting neurons with $\Delta_{n}$ values greater than the minimum $\Delta_{n}$ of the typo neurons in $t=1$ for each model. In other words, we extracted neurons that activate equally to or greater than the typo neurons in $t=1$ as typo neurons.
Figure~\ref{fig:neurons_results_t16_diff} shows the layer-wise distribution of typo neurons under this new criterion.
This shows that while typo neurons increase in the late layers of Llama 3 family and Qwen 2.5, they also increase significantly in the middle layers.
For Gemma 2, the typo neurons in the early layers decrease, while those in the late layers increase even in Figure~\ref{fig:neurons_results_t16_diff}.
This suggests that both the early and late layers are responsible for recognizing local contexts and the balance of responsibility between them can shift.

The number of typo neurons in Qwen 2.5 32B and Gemma 2 27B does not increase compared to the case of $t=1$ in \S \ref{sec:neuron_results}, while the number of typo neurons in most other models significantly increases in Figure~\ref{fig:neurons_results_t16_diff}. This suggests that typo neurons in larger models can fix typos regardless of the number of typos.

\section{Neuron Ablation for Other Models}
\label{sec:neuron_ablation_others}
\begin{table}[t]
    \small
    \centering
    \begin{tabular}{ll|rr}
        \hline \hline
         &  & \multicolumn{1}{l}{\begin{tabular}[c]{@{}l@{}}Clean\\ Dataset\end{tabular}} & \multicolumn{1}{l}{\begin{tabular}[c]{@{}l@{}}Typo\\ Dataset\end{tabular}} \\ \hline
        \multicolumn{2}{l|}{Llama 3.2 1B} & 1.00 & 0.69 \\
         & $\ominus$ Random Neurons&  0.91  & 0.61 \\
         & $\ominus$ Typo Neurons& 0.73 & 0.46 \\ \hdashline
        \multicolumn{2}{l|}{Llama 3.2 3B} & 1.00 & 0.90 \\
         & $\ominus$ Random Neurons & 0.97 & 0.89 \\
         & $\ominus$ Typo Neurons & 0.87 &0.79 \\ \hdashline
        \multicolumn{2}{l|}{Llama 3.1 8B} & 1.00 & 0.94 \\
         & $\ominus$ Random Neurons &0.99 &  0.93\\
         & $\ominus$ Typo Neurons & 0.83 & 0.80 \\ \hline
        \multicolumn{2}{l|}{Qwen 2.5 3B} & 1.00 & 0.92\\
         & $\ominus$ Random Neurons& 0.99 & 0.91 \\
         & $\ominus$ Typo Neurons& 0.84 & 0.71 \\ \hdashline
        \multicolumn{2}{l|}{Qwen 2.5 7B} & 1.00 & 0.92 \\
         & $\ominus$ Random Neurons & 0.98 & 0.92 \\
         & $\ominus$ Typo Neurons & 0.86 & 0.80 \\ \hdashline
         \multicolumn{2}{l|}{Qwen 2.5 14B} & 1.00 & 0.95 \\
         & $\ominus$ Random Heads & 0.99 & 0.94 \\
         & $\ominus$ Typo Heads & 0.92 & 0.82 \\ \hdashline
        \multicolumn{2}{l|}{Qwen 2.5 32B} & 1.00 & 0.96 \\
         & $\ominus$ Random Neurons & 0.99 & 0.96 \\
         & $\ominus$ Typo Neurons & 0.93 & 0.85 \\          
         \hline \hline
    \end{tabular}
    \caption{Accuracy of the word identification task with neuron ablation on clean and typo datasets. 
    ``$\ominus$ Random Neurons'' and ``$\ominus$ Typo Neurons'' indicate the performance by ablating random and typo neurons, respectively.
    }
    \label{tab:deactivate_result_others}
\end{table}
In \S \ref{sec:neuron_ablation}, we reported the results for Gemma 2. Here, we examined the ablation study for typo neurons in the Llama 3 family and Qwen 2.5. 
 
Table~\ref{tab:deactivate_result_others} shows that the results of the ablation study are consistent, while there were differences in typo neuron distributions across models. 
In all models, ablating random neurons did not reduce accuracy on the typo dataset. In contrast, ablating typo neurons led to a drop in accuracy on both the clean and typo datasets.
This indicates that typo neurons may not exclusively act on typos but could also play a crucial role in processing general grammatical or morphological features, regardless of the model. 

\section{Typo Heads for Many Typos}
\label{sec:t_16_heads}
\begin{figure*}[t]
    \centering
    \includegraphics[width=\linewidth]{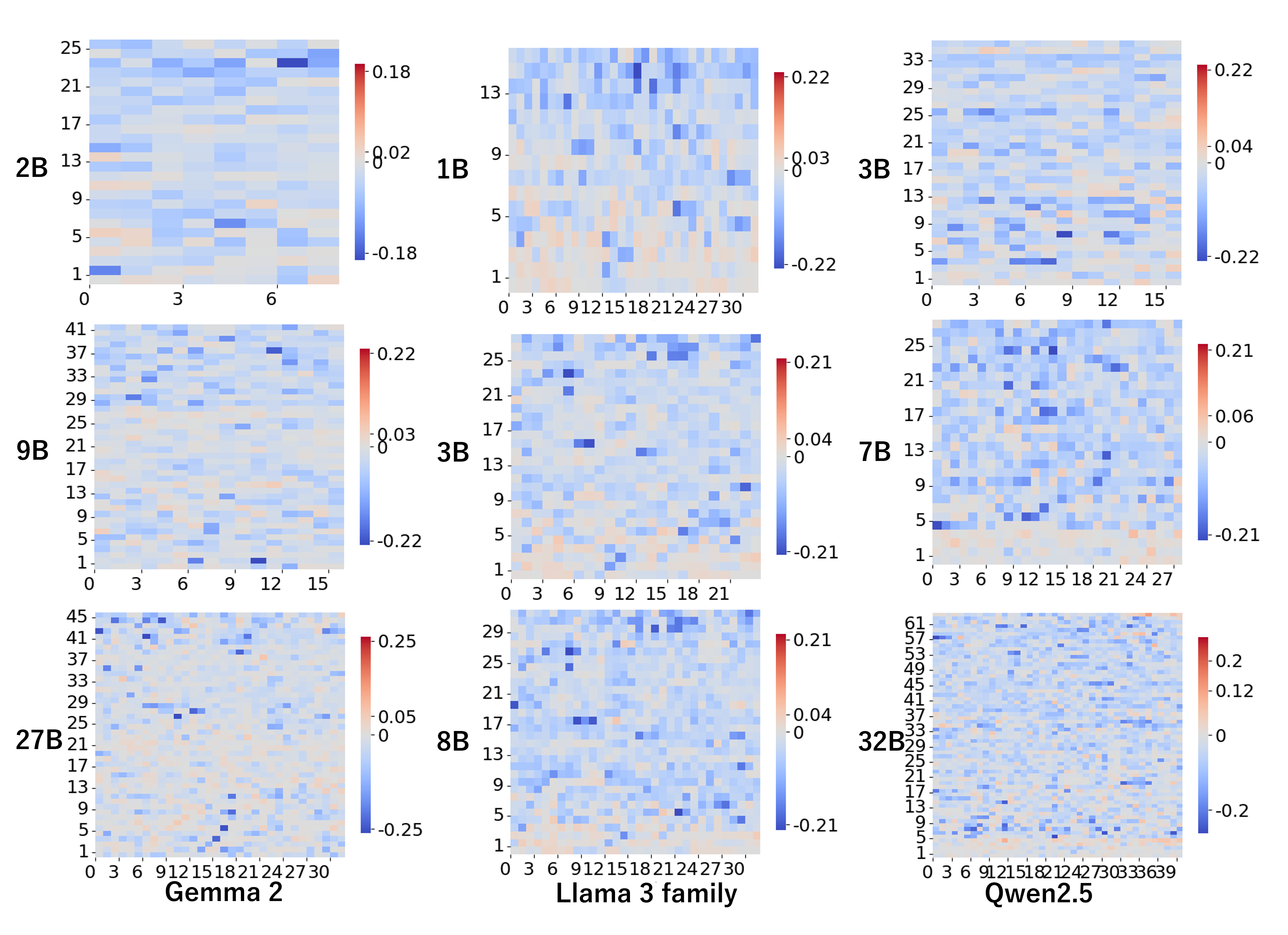}
    \caption{Distribution of ${\Delta}_{h}$ for each model with $t=16$. 
    The heat map colors are centered around 0, and the tick mark closest to 0 on the positive side of the heat bar represents the maximum ${\Delta}_{h}$. The left figures are for Gemma 2, the center figures are for Llama 3 family and the right figures are for Qwen 2.5.}
    \label{fig:head_delta_t16}
\end{figure*}

\begin{table*}[t]
    \small
    \centering
    \begin{tabular}{l|rrr:rrr:rrrr}
        \hline \hline
         & \multicolumn{3}{c:}{Gemma 2} & \multicolumn{2}{c}{Llama 3.2} & \multicolumn{1}{c:}{Llama 3.1} & \multicolumn{4}{c}{Qwen 2.5} \\ 
         & \multicolumn{1}{c}{2B} & \multicolumn{1}{c}{9B} & \multicolumn{1}{c}{27B} & \multicolumn{1}{c}{1B} & \multicolumn{1}{c}{3B} & \multicolumn{1}{c}{8B} & \multicolumn{1}{c}{3B} & \multicolumn{1}{c}{7B} & \multicolumn{1}{c}{14B} & \multicolumn{1}{c}{32B} \\ \hline
        Average & -0.0295 & -0.0276 & -0.0221 & -0.0330 & -0.0295 & -0.0368 & -0.0347 & -0.0401 & -0.0343 & -0.0369 \\
        \begin{tabular}[c]{@{}l@{}}Standard \\ Deviation\end{tabular} & 0.0317 & 0.0335 & 0.0394 & 0.0442 & 0.0383 & 0.0398 & 0.0557 & 0.0434 & 0.0420 & 0.0452 \\ \hline \hline
    \end{tabular}
    \caption{The average and standard deviation of $\Delta_{h}$ with $t=16$.}
    \label{tab:head_ave_sd_t16}
\end{table*}

Similar to Appendix~\ref{sec:t_16_neurons}, while \S \ref{sec:head_results} reported for $t=1$, here we describe the behavior of typo heads under the $t=16$ setting.

Table~\ref{tab:head_ave_sd_t16} shows that $\Delta_{h}$ shifts significantly in the negative direction at $t=16$ compared to $t=1$. the minimum values in Figure~\ref{fig:head_delta_t16} also shows this transition.
Additionally, the increase in dark blue areas in Figure~\ref{fig:head_delta_t16} indicates that more heads respond relatively strongly. However, the difference between $t=1$ and $t=16$ for typo heads is smaller than for typo neurons.

\section{Typo Heads for Qwen 2.5 14B}
\label{sec:qwen14b_heads}
\begin{figure}[t]
    \centering
    \includegraphics[width=\linewidth]{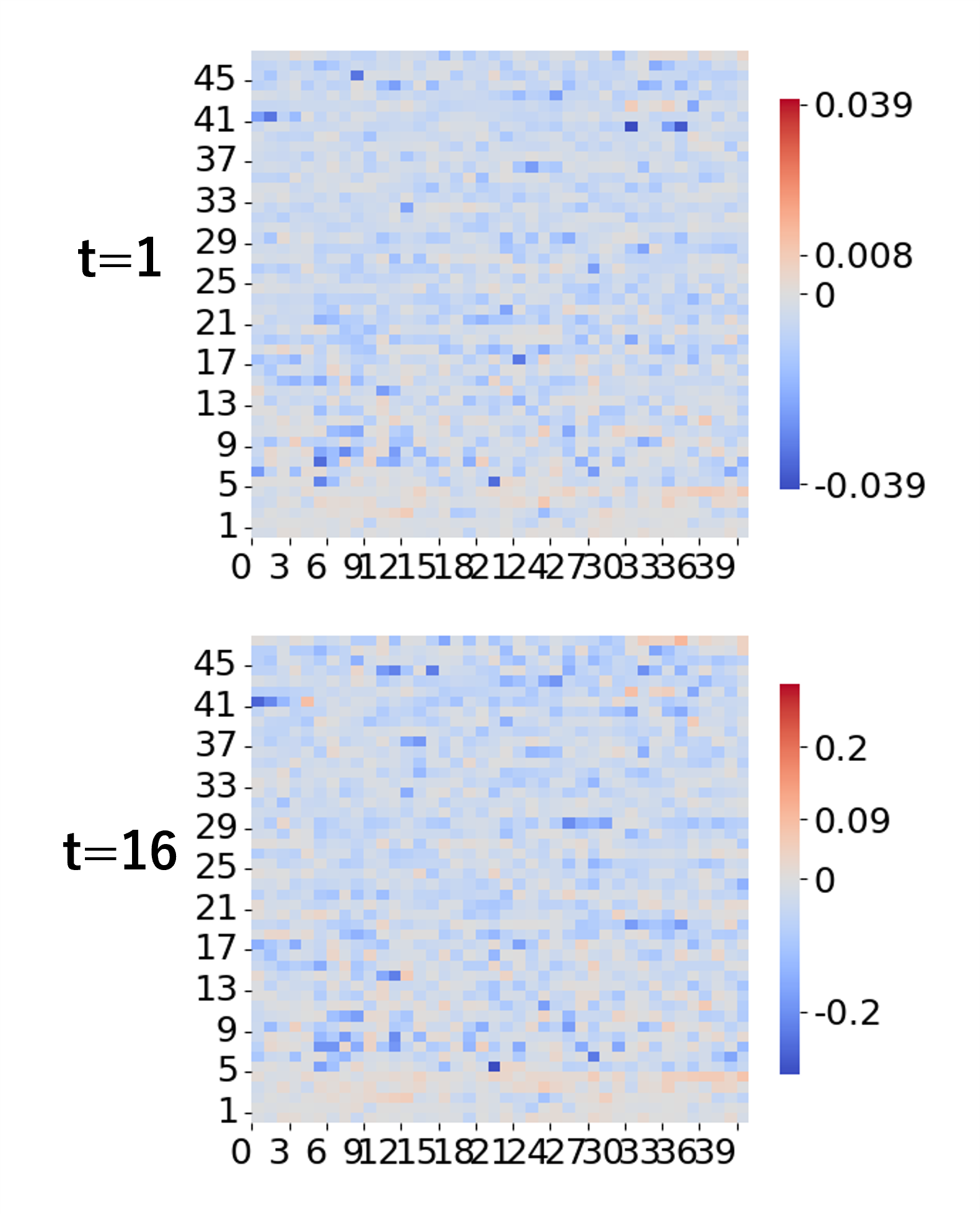}
    \caption{Distribution of ${\Delta}_{h}$ for Qwen 2.5 14B. 
    The heat map colors are centered around 0, and the tick mark closest to 0 on the positive side of the heat bar represents the maximum ${\Delta}_{h}$.}
    \label{fig:head_delta_qwen14}
\end{figure}

Figure~\ref{fig:head_delta_qwen14} shows the distribution of $\Delta_{h}$ for Qwen 2.5 14B, which was not included in \S \ref{sec:head_results} and Appendix~\ref{sec:t_16_heads} due to space constraints.
The results are consistent with those of other models and model sizes, as the initial layers contain fewer typo heads, and the distribution of typo heads is sparser than in smaller models.

\section{Head Ablation for Other Models}
\label{sec:head_ablation_others}

\begin{table}[t]
    \small
    \centering
    \begin{tabular}{ll|rr}
        \hline \hline
         &  & \multicolumn{1}{l}{\begin{tabular}[c]{@{}l@{}}Clean\\ Dataset\end{tabular}} & \multicolumn{1}{l}{\begin{tabular}[c]{@{}l@{}}Typo\\ Dataset\end{tabular}} \\ \hline
        \multicolumn{2}{l|}{Llama 3.2 1B} & 1.00 & 0.69 \\
         & $\ominus$ Random Heads& 0.07 & 0.04 \\
         & $\ominus$ Typo Heads& 0.00 & 0.00 \\ \hdashline
        \multicolumn{2}{l|}{Llama 3.2 3B} & 1.00 & 0.90 \\
         & $\ominus$ Random Heads & 0.10 & 0.10 \\
         & $\ominus$ Typo Heads & 0.18 & 0.17 \\ \hdashline
        \multicolumn{2}{l|}{Llama 3.1 8B} & 1.00 & 0.94 \\
         & $\ominus$ Random Heads & 0.09  & 0.08 \\
         & $\ominus$ Typo Heads & 0.10 & 0.09 \\ \hline
        \multicolumn{2}{l|}{Qwen 2.5 3B} & 1.00 & 0.92 \\
         & $\ominus$ Random Heads& 0.97 & 0.88 \\
         & $\ominus$ Typo Heads& 0.46 & 0.41 \\ \hdashline
        \multicolumn{2}{l|}{Qwen 2.5 7B} & 1.00 & 0.92 \\
         & $\ominus$ Random Heads & 0.55 & 0.53 \\
         & $\ominus$ Typo Heads & 0.39 & 0.37 \\ \hdashline
         \multicolumn{2}{l|}{Qwen 2.5 14B} & 1.00 & 0.95 \\
         & $\ominus$ Random Heads & 0.09 & 0.09 \\
         & $\ominus$ Typo Heads & 0.13 & 0.12 \\ \hdashline
        \multicolumn{2}{l|}{Qwen 2.5 32B} & 1.00 & 0.96 \\
         & $\ominus$ Random Heads & 0.18 & 0.16 \\
         & $\ominus$ Typo Heads & 0.15 & 0.15 \\          
         \hline \hline
    \end{tabular}
    \caption{Accuracy of the word identification task with head ablation on clean and typo datasets. 
    ``$\ominus$ Random Heads'' and ``$\ominus$ Typo Heads'' indicate the performance by ablating random and typo heads, respectively.
    }
    \label{tab:head_deactivate_result_others}
\end{table}

Similar to Appendix~\ref{sec:neuron_ablation_others}, we examined the ablation study for typo heads in the Llama 3 family and Qwen 2.5. 

In Table~\ref{tab:head_deactivate_result_others}, both ablations significantly degraded the model's capability in the Llama 3 family, Qwen 2.5 14B and Qwen 2.5 32B, making it difficult to determine the importance of typo heads.
In contrast, in Qwen 2.5 3B and Qwen 2.5 7B, the ablation of typo heads decreases accuracy more than the ablation of random heads. Compared to \S \ref{sec:head_ablation}, where ablation of typo heads in the 9B model had little impact on accuracy, this suggests that typo heads remain important even in the middle model in Qwen 2.5, which has few typo neurons and typo heads in the early layers.

\section{Visualization of Typo Heads.}

\begin{figure*}[t]
    \centering
    \includegraphics[width=15.5cm]{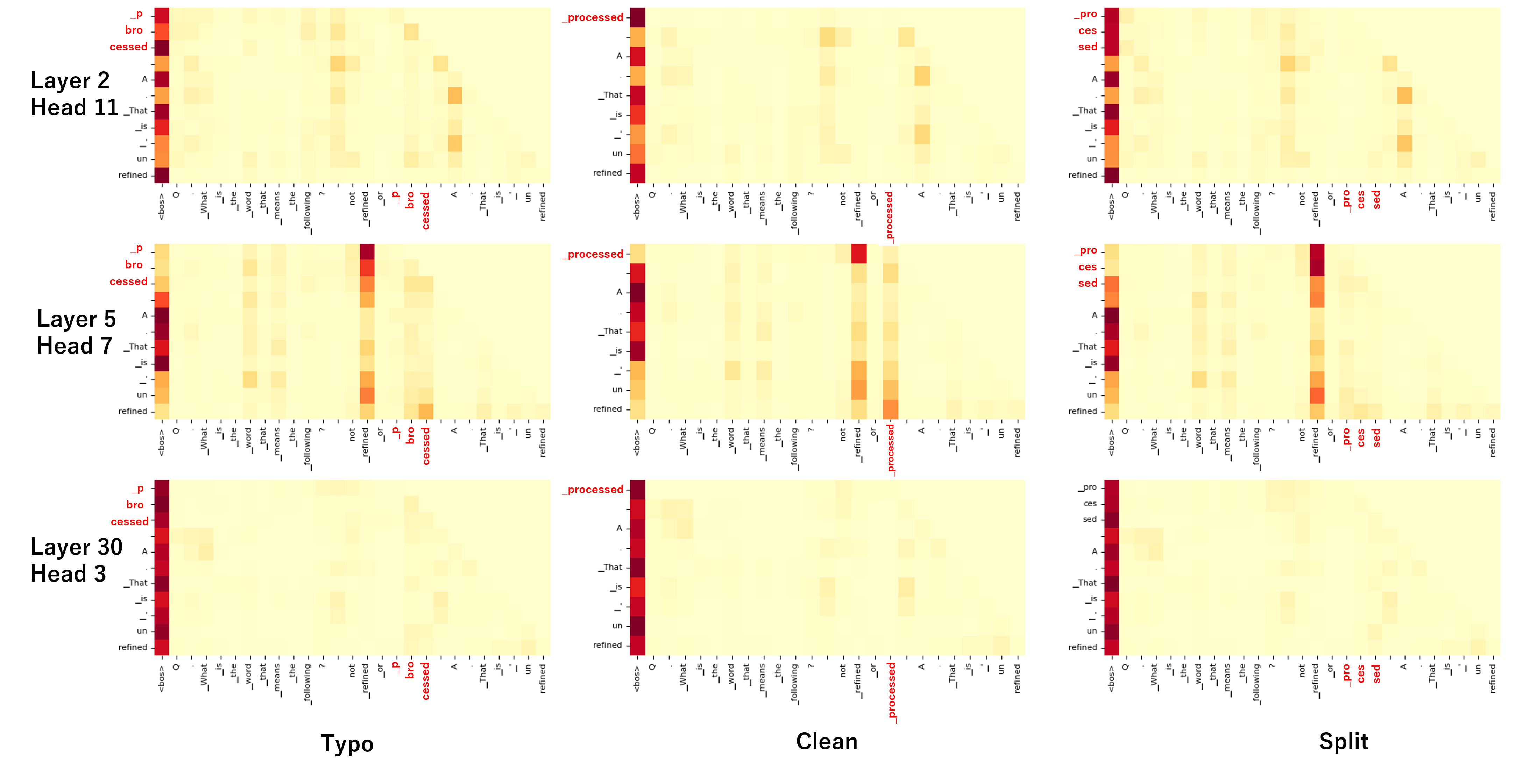}
    \caption{Visualization of typo heads in the 9B model. The word definition in the clean input is ``not refined or processed,'' and the correct answer is ``unrefined''. The word ``processed'' was changed with a typo to ``pbrocessed.'' 
}
    \label{fig:head_case_study}
\end{figure*}

Figure~\ref{fig:head_case_study} shows the attention maps for each input, using the top 1.5\% of heads with the highest absolute value of ${\Delta}_{h}$ scores in Gemma 2 9B as typo heads.

The typo head in Layer 2 Head 11 recognizes sentence boundaries.
This head is not a head that contributes to typo-fixing but is damaged by typos. Our method has a limitation in that it cannot distinguish between heads that contribute to typo-fixing and those that are damaged by typos.
The typo head in Layer 5 Head 7 responds to semantic connections and fixes typos by leveraging synonyms. 
This is a typical typo-fixing mechanism of early middle layers described above, which is a recognition of global contexts.
The typo head in Layer 30 Head 3 fixes typos by recognizing local contexts.
Additionally, most typo heads strongly attend to '\verb|<bos>|'.

\section{Future Work}
This paper focuses on the investigation of typo-related inner workings. 
Therefore, we do not provide any methods to improve LLM's robustness against typos.
However, our findings imply how to create more robust LLMs against typos.

Our findings indicate that typo neurons in the early or late layers of Transformer-based LLMs fix typos with local contexts, while typo neurons in the middle layers fix typos with global contexts. 
The model's robustness against typos may enhanced by a mechanism that gives more importance to nearby tokens in the early and late layers and to distant tokens in the middle layers.

Furthermore, the results of the ablation study show that typo-fixing is related to general grammatical or morphological recognition, which suggests that methods for improving general contextual recognition could contribute to typo robustness. 
For example, a potential research direction could be investigating how additional training on tasks such as grammatical error correction or determining whether a given subword is part of a specific word affects robustness against typos.

\end{document}